\ifx\XeTeXrevision\undefined \pdfoutput=1 \fi  % arXiv: build with pdflatex (figures are PDF)
\documentclass[11pt]{article}
\usepackage[preprint]{acl}   % named version: review mode prints "Anonymous ACL submission" instead
\usepackage{times}
\usepackage{latexsym}
\usepackage{booktabs}
\usepackage{amssymb}
\usepackage{graphicx}
\usepackage{inconsolata}

\usepackage[T1]{fontenc}
\usepackage[utf8]{inputenc}
\usepackage{microtype}
\graphicspath{{fig/}}
\hypersetup{pdftitle={Type-Safe Is Not Error-Free: Typed Decision Models Follow the Option Name, Not the Definition Bound to It},
            pdfauthor={Yu Sun, Junhao Xu, Jiajia Shi, Zijin Yang}}
\newcommand{\m}[1]{\csname pn#1\endcsname}       % \m{flipNY} -> 76.92
\newcommand{\didn}[1]{\csname pn#1\endcsname}    % same, for signed point differences
\expandafter\def\csname pnapiAucANY\endcsname{81.5}
\expandafter\def\csname pnapiAucSNY\endcsname{58.1}
\expandafter\def\csname pnapiBalANY\endcsname{71.3}
\expandafter\def\csname pnapiBalSNY\endcsname{51.6}
\expandafter\def\csname pnapiCiNYhi\endcsname{33.33}
\expandafter\def\csname pnapiCiNYlo\endcsname{27.58}
\expandafter\def\csname pnapiDecNeut\endcsname{1.88}
\expandafter\def\csname pnapiDecPolar\endcsname{11.06}
\expandafter\def\csname pnapiDecVocab\endcsname{32.21}
\expandafter\def\csname pnapiDidNY\endcsname{30.42}
\expandafter\def\csname pnapiFlipAB\endcsname{1.67}
\expandafter\def\csname pnapiFlipAP\endcsname{15.17}
\expandafter\def\csname pnapiFlipFT\endcsname{31.92}
\expandafter\def\csname pnapiFlipNP\endcsname{4.17}
\expandafter\def\csname pnapiFlipNY\endcsname{32.50}
\expandafter\def\csname pnapiFlipRA\endcsname{13.83}
\expandafter\def\csname pnapiFlipZO\endcsname{2.08}
\expandafter\def\csname pnapiFloor\endcsname{1.33}
\expandafter\def\csname pnapiFloorPairs\endcsname{2}
\expandafter\def\csname pnapiFloorRatio\endcsname{24}
\expandafter\def\csname pnapiIdProbMax\endcsname{36.33}
\expandafter\def\csname pnapiRatioNY\endcsname{2.4}
\expandafter\def\csname pnaucAAB\endcsname{94.2}
\expandafter\def\csname pnaucAAP\endcsname{94.5}
\expandafter\def\csname pnaucAFT\endcsname{96.2}
\expandafter\def\csname pnaucANP\endcsname{82.6}
\expandafter\def\csname pnaucANY\endcsname{93.8}
\expandafter\def\csname pnaucARA\endcsname{90.8}
\expandafter\def\csname pnaucAZO\endcsname{94.2}
\expandafter\def\csname pnaucSAB\endcsname{94.1}
\expandafter\def\csname pnaucSAP\endcsname{54.3}
\expandafter\def\csname pnaucSFT\endcsname{57.7}
\expandafter\def\csname pnaucSNP\endcsname{75.8}
\expandafter\def\csname pnaucSNY\endcsname{23.2}
\expandafter\def\csname pnaucSRA\endcsname{63.6}
\expandafter\def\csname pnaucSZO\endcsname{92.4}
\expandafter\def\csname pnaucUpSFT\endcsname{3.0}
\expandafter\def\csname pnaucUpSNY\endcsname{8.6}
\expandafter\def\csname pnbalAAB\endcsname{84.2}
\expandafter\def\csname pnbalAAP\endcsname{85.9}
\expandafter\def\csname pnbalAFT\endcsname{90.1}
\expandafter\def\csname pnbalANP\endcsname{75.3}
\expandafter\def\csname pnbalANY\endcsname{87.2}
\expandafter\def\csname pnbalARA\endcsname{79.1}
\expandafter\def\csname pnbalAZO\endcsname{83.7}
\expandafter\def\csname pnbalSAB\endcsname{82.2}
\expandafter\def\csname pnbalSAP\endcsname{53.5}
\expandafter\def\csname pnbalSFT\endcsname{48.6}
\expandafter\def\csname pnbalSNP\endcsname{67.3}
\expandafter\def\csname pnbalSNY\endcsname{28.4}
\expandafter\def\csname pnbalSRA\endcsname{66.2}
\expandafter\def\csname pnbalSZO\endcsname{83.2}
\expandafter\def\csname pnciAPZohi\endcsname{44.17}
\expandafter\def\csname pnciAPZolo\endcsname{38.17}
\expandafter\def\csname pnciFTZohi\endcsname{46.08}
\expandafter\def\csname pnciFTZolo\endcsname{40.25}
\expandafter\def\csname pnciNPZohi\endcsname{44.83}
\expandafter\def\csname pnciNPZolo\endcsname{38.75}
\expandafter\def\csname pnciNYAbhi\endcsname{73.50}
\expandafter\def\csname pnciNYAblo\endcsname{68.16}
\expandafter\def\csname pnciNYZohi\endcsname{73.08}
\expandafter\def\csname pnciNYZolo\endcsname{67.58}
\expandafter\def\csname pnciRAZohi\endcsname{42.25}
\expandafter\def\csname pnciRAZolo\endcsname{36.33}
\expandafter\def\csname pnciUpNYhi\endcsname{81.33}
\expandafter\def\csname pnciUpNYlo\endcsname{73.00}
\expandafter\def\csname pncompAucMax\endcsname{98.7}
\expandafter\def\csname pncompAucMin\endcsname{91.4}
\expandafter\def\csname pncompBalMax\endcsname{93.1}
\expandafter\def\csname pncompBalMin\endcsname{78.6}
\expandafter\def\csname pndecNeut\endcsname{6.25}
\expandafter\def\csname pndecNeutHi\endcsname{11.3}
\expandafter\def\csname pndecNeutLo\endcsname{2.7}
\expandafter\def\csname pndecPolar\endcsname{47.25}
\expandafter\def\csname pndecPolarHi\endcsname{92.0}
\expandafter\def\csname pndecPolarLo\endcsname{12.3}
\expandafter\def\csname pndecVocab\endcsname{63.29}
\expandafter\def\csname pndecVocabHi\endcsname{92.7}
\expandafter\def\csname pndecVocabLo\endcsname{29.0}
\expandafter\def\csname pndidAPZo\endcsname{41.25}
\expandafter\def\csname pndidFTZo\endcsname{43.17}
\expandafter\def\csname pndidNPZo\endcsname{41.75}
\expandafter\def\csname pndidNYAb\endcsname{70.92}
\expandafter\def\csname pndidNYZo\endcsname{70.42}
\expandafter\def\csname pndidRAZo\endcsname{39.25}
\expandafter\def\csname pndidUpNY\endcsname{77.17}
\expandafter\def\csname pnebAucAB\endcsname{36.0}
\expandafter\def\csname pnebAucFT\endcsname{78.1}
\expandafter\def\csname pnebAucFTs\endcsname{51.4}
\expandafter\def\csname pnebAucIC\endcsname{74.3}
\expandafter\def\csname pnebAucNY\endcsname{78.6}
\expandafter\def\csname pnebAucOPT\endcsname{42.2}
\expandafter\def\csname pnebAucZO\endcsname{76.3}
\expandafter\def\csname pnebBalAB\endcsname{34.0}
\expandafter\def\csname pnebBalFT\endcsname{61.7}
\expandafter\def\csname pnebBalFTs\endcsname{57.7}
\expandafter\def\csname pnebBalIC\endcsname{57.0}
\expandafter\def\csname pnebBalNY\endcsname{73.6}
\expandafter\def\csname pnebBalOPT\endcsname{41.2}
\expandafter\def\csname pnebBalZO\endcsname{57.7}
\expandafter\def\csname pnflipAB\endcsname{6.00}
\expandafter\def\csname pnflipAP\endcsname{47.75}
\expandafter\def\csname pnflipFT\endcsname{49.67}
\expandafter\def\csname pnflipNP\endcsname{48.25}
\expandafter\def\csname pnflipNY\endcsname{76.92}
\expandafter\def\csname pnflipRA\endcsname{45.75}
\expandafter\def\csname pnflipUpNY\endcsname{87.67}
\expandafter\def\csname pnflipZO\endcsname{6.50}
\expandafter\def\csname pngoldZeroShare\endcsname{14.1}
\expandafter\def\csname pnhlAucA\endcsname{94}
\expandafter\def\csname pnhlAucS\endcsname{23}
\expandafter\def\csname pnhlDid\endcsname{70.4}
\expandafter\def\csname pnhlInv\endcsname{77}
\expandafter\def\csname pnincrPolar\endcsname{41.00}
\expandafter\def\csname pnincrVocab\endcsname{16.04}
\expandafter\def\csname pnkAccNeut\endcsname{27.8}
\expandafter\def\csname pnkAccNeutFour\endcsname{48.5}
\expandafter\def\csname pnkAccNum\endcsname{25.9}
\expandafter\def\csname pnkAccNumFour\endcsname{45.6}
\expandafter\def\csname pnkAccOrig\endcsname{56.4}
\expandafter\def\csname pnkAccOrigFour\endcsname{49.5}
\expandafter\def\csname pnkAccOrigSix\endcsname{16.7}
\expandafter\def\csname pnkAccOrigSixteen\endcsname{64.6}
\expandafter\def\csname pnkAccRot\endcsname{15.5}
\expandafter\def\csname pnkAccRotFour\endcsname{38.7}
\expandafter\def\csname pnkCards\endcsname{3, 4, 5, 6, 16}
\expandafter\def\csname pnkChgNeut\endcsname{52.42}
\expandafter\def\csname pnkChgNum\endcsname{55.78}
\expandafter\def\csname pnkChgRot\endcsname{79.65}
\expandafter\def\csname pnkFollowsRot\endcsname{24.0}
\expandafter\def\csname pnkGap\endcsname{27.2}
\expandafter\def\csname pnkNeutSixteen\endcsname{14.1}
\expandafter\def\csname pnnApi\endcsname{1200}
\expandafter\def\csname pnnChoice\endcsname{683}
\expandafter\def\csname pnnEb\endcsname{1668}
\expandafter\def\csname pnnFloor\endcsname{300}
\expandafter\def\csname pnnOj\endcsname{1800}
\expandafter\def\csname pnnOpq\endcsname{15}
\expandafter\def\csname pnnOpqApi\endcsname{5}
\expandafter\def\csname pnnPerPred\endcsname{300}
\expandafter\def\csname pnnPoolsKept\endcsname{5}
\expandafter\def\csname pnnPred\endcsname{4}
\expandafter\def\csname pnnPrim\endcsname{1200}
\expandafter\def\csname pnnRec\endcsname{1200}
\expandafter\def\csname pnnUp\endcsname{600}
\expandafter\def\csname pnojDecNeut\endcsname{14.80}
\expandafter\def\csname pnojDecPolar\endcsname{18.56}
\expandafter\def\csname pnojDecVocab\endcsname{33.66}
\expandafter\def\csname pnojFlipAB\endcsname{26.00}
\expandafter\def\csname pnojFlipAP\endcsname{24.06}
\expandafter\def\csname pnojFlipFT\endcsname{47.83}
\expandafter\def\csname pnojFlipNP\endcsname{23.44}
\expandafter\def\csname pnojFlipNY\endcsname{19.50}
\expandafter\def\csname pnojFlipRA\endcsname{8.17}
\expandafter\def\csname pnojFlipZO\endcsname{3.61}
\expandafter\def\csname pnojRatioNY\endcsname{4.1}
\expandafter\def\csname pnopqBalApiNeut\endcsname{71.5}
\expandafter\def\csname pnopqBalApiOpq\endcsname{71.5}
\expandafter\def\csname pnopqBalApiPolar\endcsname{70.9}
\expandafter\def\csname pnopqBalApiVocab\endcsname{71.1}
\expandafter\def\csname pnopqBalMkNeut\endcsname{87.1}
\expandafter\def\csname pnopqBalMkOpq\endcsname{86.2}
\expandafter\def\csname pnopqBalMkPolar\endcsname{79.4}
\expandafter\def\csname pnopqBalMkVocab\endcsname{89.6}
\expandafter\def\csname pnopqBalSpNeut\endcsname{63.9}
\expandafter\def\csname pnopqBalSpOpq\endcsname{60.6}
\expandafter\def\csname pnopqBalSpPolar\endcsname{59.8}
\expandafter\def\csname pnopqBalSpVocab\endcsname{60.1}
\expandafter\def\csname pnopqCiApihi\endcsname{0.67}
\expandafter\def\csname pnopqCiApilo\endcsname{-0.38}
\expandafter\def\csname pnopqCiMkhi\endcsname{-0.18}
\expandafter\def\csname pnopqCiMklo\endcsname{-1.98}
\expandafter\def\csname pnopqCiSphi\endcsname{-2.26}
\expandafter\def\csname pnopqCiSplo\endcsname{-4.02}
\expandafter\def\csname pnopqDbalMax\endcsname{3.3}
\expandafter\def\csname pnopqDidApi\endcsname{0.14}
\expandafter\def\csname pnopqDidMk\endcsname{-1.09}
\expandafter\def\csname pnopqDidSp\endcsname{-3.16}
\expandafter\def\csname pnopqFlipApiNeut\endcsname{1.88}
\expandafter\def\csname pnopqFlipApiOpq\endcsname{2.02}
\expandafter\def\csname pnopqFlipApiPolar\endcsname{11.06}
\expandafter\def\csname pnopqFlipApiVocab\endcsname{32.21}
\expandafter\def\csname pnopqFlipMkNeut\endcsname{7.94}
\expandafter\def\csname pnopqFlipMkOpq\endcsname{6.86}
\expandafter\def\csname pnopqFlipMkPolar\endcsname{50.22}
\expandafter\def\csname pnopqFlipMkVocab\endcsname{70.72}
\expandafter\def\csname pnopqFlipSpNeut\endcsname{14.83}
\expandafter\def\csname pnopqFlipSpOpq\endcsname{11.67}
\expandafter\def\csname pnopqFlipSpPolar\endcsname{18.57}
\expandafter\def\csname pnopqFlipSpVocab\endcsname{33.75}
\expandafter\def\csname pnopqGapMax\endcsname{3.16}
\expandafter\def\csname pnopqLen\endcsname{5}
\expandafter\def\csname pnopqMaxApi\endcsname{2.42}
\expandafter\def\csname pnopqMaxMk\endcsname{13.28}
\expandafter\def\csname pnopqMaxSp\endcsname{31.61}
\expandafter\def\csname pnopqMinApi\endcsname{1.67}
\expandafter\def\csname pnopqMinMk\endcsname{3.22}
\expandafter\def\csname pnopqMinSp\endcsname{2.00}
\expandafter\def\csname pnopqSdApi\endcsname{0.34}
\expandafter\def\csname pnopqSdMk\endcsname{2.46}
\expandafter\def\csname pnopqSdSp\endcsname{8.66}
\expandafter\def\csname pnopqSpNy\endcsname{19.50}
\expandafter\def\csname pnopqVocabMax\endcsname{70.72}
\expandafter\def\csname pnopqVocabMin\endcsname{32.21}
\expandafter\def\csname pnpoolFlipAB\endcsname{8.06}
\expandafter\def\csname pnpoolFlipAP\endcsname{51.00}
\expandafter\def\csname pnpoolFlipFT\endcsname{60.94}
\expandafter\def\csname pnpoolFlipNP\endcsname{50.50}
\expandafter\def\csname pnpoolFlipNY\endcsname{80.50}
\expandafter\def\csname pnpoolFlipRA\endcsname{49.17}
\expandafter\def\csname pnpoolFlipZO\endcsname{7.83}
\expandafter\def\csname pnptCtlMax\endcsname{11.3}
\expandafter\def\csname pnptNYmin\endcsname{56.7}
\expandafter\def\csname pnptOtherMax\endcsname{92.0}
\expandafter\def\csname pnptOtherMin\endcsname{12.3}
\expandafter\def\csname pnptRatioMin\endcsname{7.4}
\expandafter\def\csname pnyesRate\endcsname{57.5}
\expandafter\def\csname pnyesRateApi\endcsname{57.5}
\newcommand{\opt}[1]{\textsf{#1}}          % an option name, as the model sees it

\usepackage{siunitx}

\title{Type-Safe Is Not Error-Free: Typed Decision Models\\Follow the Option Name, Not the Definition Bound to It}
\author{
  Yu Sun\thanks{\ Equal contribution.} \\
  National University of Singapore \\
  \texttt{sun.yu@u.nus.edu} \\\And
  Junhao Xu\footnotemark[1] \\
  Fudan University \\
  \texttt{junhaoxu23@m.fudan.edu.cn} \\\AND
  Jiajia Shi \\
  Fudan University \\
  \texttt{shijj25@m.fudan.edu.cn} \\\And
  Zijin Yang \\
  University of Science and Technology of China \\
  \texttt{bsmhmmlf@mail.ustc.edu.cn} \\
}
\date{}

\begin{document}
\maketitle

\begin{abstract}
Typed decision models return structured results, but output-type correctness alone does not ensure that decisions follow explicit option definitions.
Each option pairs a name with a definition that defines its intended meaning; the name, however, can provide a competing semantic cue.
We study this conflict in Jev and two open-weight models by changing only the name-definition mapping, leaving the question, state, and the names and definition texts themselves unchanged.
We measure decision flips at the level of the selected definition, rather than the returned name.
On \m{nPrim} decision tasks with task-specific definitions, decision-flip rates are up to \m{hlDid} pp higher with yes/no names than with the 0/1 control.
This gap holds across all \m{nPred} binary decision rules.
With yes/no names, reassignment also lowers their mean AUC from \m{aucANY}\% to a below-chance \m{aucSNY}\%.
In the binary evaluations, random strings used as option names yield mean flip rates close to those of neutral controls across all three models, with comparable balanced accuracy before reassignment.
Together, these results support option-name polarity as a contributor to decision instability beyond reassignment alone.
The type-error rate remains 0\% throughout, showing that type-correct outputs can still fail to follow explicit option definitions.
\end{abstract}

\section{Introduction}
Typed decision models, such as Jev, return decisions rather than free-form text.
Given a typed question, a state, and a declared set of options, such a model returns a probability distribution over that set.
Here, type safety refers to the guarantee that the model assigns zero probability to any candidate outside the declared option set. In Jev, this property holds by construction because scores for candidates outside the set are masked before normalization.
The blog post introducing Jev therefore presents the model's 0\% type-error rate
as a structural guarantee rather than an empirical measurement.\footnote{
\url{https://typesafe.ai/blog/introducing-system-one-models-and-jev}}
Type safety restricts which options a model can return, but does not ensure that it selects among them according to their explicit definitions.

Each option pairs a short name with a definition that defines its intended meaning for the task. Names such as yes/no, however, carry conventional meanings that may conflict with these definitions.
Just as a cat named ``Dog'' remains a cat, an option's intended meaning
is defined by its definition, not by the conventional meaning of its
name. We therefore ask whether models follow explicit option definitions
when option names provide competing semantic cues.

To study this conflict, we keep the question, the state, the option
names, and the definition texts fixed, and change only how names are
paired with definitions. We track changes in the selected definition
rather than in the returned name. Under a binary swap, a model that
follows the definitions should preserve its definition-level choice
and therefore return a different name. Returning the same name instead
selects a different definition. Each input states the name--definition
assignments explicitly, so the model need not infer them from demonstrations.

Prior work has shown that language models are sensitive to prompt format,
option identifiers, and label words
\citep{sclar2024formatspread,zheng2024mcq,liusie2023wordbias}, and that
smaller models tend to follow the semantic priors of label words rather than
override them \citep{wei2023flipped}. Recent studies further show that
schema-valid outputs can still be semantically incorrect, and that a schema
can take precedence over an explicit instruction
\citep{li2026json,singh2026sob,usman2026phantomfill,yourprompt2026}.
Our work asks a narrower question: when an option's name and its definition
suggest different meanings, which one determines the model's decision?
We study this question on models whose outputs are type-safe by construction,
so the observed failures cannot be attributed to invalid outputs.

Under this setting, the effect of polar names goes beyond a drop in accuracy.
For yes/no names, reassignment lowers the AUC of one model below chance,
meaning that its scores rank negative instances above positive ones more
often than the reverse. Because AUC is independent of the decision threshold,
this behavior cannot be corrected by threshold adjustment.

\noindent Our main findings are as follows.

\begin{enumerate}
\item Option names can override explicit definitions.
Polar-name swaps affect decisions across every tested predicate and can reverse score rankings, rather than merely change individual answers.

\item Polarity contributes to decision instability.
Familiar polar names, including interface defaults, are most affected, while random strings behave like neutral identifiers with comparable accuracy.

\item Name dependence extends beyond binary decision tasks and a single model family.
Neutral renaming changes non-binary decisions, and sensitivity to names recurs across model families, with effect sizes varying with readout geometry.
\end{enumerate}

%%%%%%%%%%%%%%分section了

\section{The scope of the type-safety guarantee}
\label{sec:lemma}

This section introduces the notation for the models we study and states what their type-safety
guarantee establishes.

\paragraph{Typed decision models.} A typed decision model receives a state $s$ and a question with
instructions $q$ and $k$ options $O=(o_1,\dots,o_k)$. The question's criteria map each option name
$n_i$, such as \opt{no} or \opt{yes}, to a definition $d_i$ of what the option means; we write
$o_i=(n_i,d_i)$. The model assigns a score $z_i$ to each option, returns
$p=\mathrm{softmax}(z)$ over $O$, and answers with the most probable option. We study Jev
\citep{typesafe} and two Jev-like models with open weights, Laya \citep{convai2026laya} and
Open-Jev \citep{kotoba2026openjev}. Jev supports three question types, Choice, Score and Noul; the
setting above is the Choice type, which is the one we study. Jev is a closed-source model accessed
through an API, so we observe only $p$. Laya is built on ModernBERT-large
\citep{warner2024modernbert} and Open-Jev on DeBERTa-v3-large \citep{he2023debertav3}. In both,
each option enters the input as \texttt{<name>: <definition>} together with the state and the
instructions, so the name and the definition are both visible to the model. They differ in how
$z_i$ is computed. Laya uses the contextual embedding of a \texttt{[MASK]} marker token placed at
the option. Open-Jev uses the average over all tokens of the option's text, covering both the name
and the definition. Section~\ref{sec:r6} discusses this difference as one possible reason why the
option name affects the two models to different degrees.

\paragraph{Type safety.} Because $p$ is a distribution over $O$ and the answer is its $\arg\max$,
the answer is always an element of $O$, for any input and any weights. In the open-weight models
this holds by construction, since scores of candidates outside $O$ are masked before normalization.
For Jev it is the documented behavior: its documentation states that the model never makes type
errors \citep{typesafe}.

\paragraph{What type safety does not cover.} Type safety constrains which options can receive
probability, but not how probability is distributed among them. This has two consequences for
evaluation. First, a 0\% type-error rate says nothing about decision quality, since it follows from
how the output is constructed rather than from what the model has learned. Second, the guarantee
holds equally in every condition of this paper, including those that invert the decision; a type
check therefore cannot detect the failure we study. What type safety leaves open is whether the
model interprets the options as intended: whether it chooses an option because its definition
matches the state, or because of its name.

\section{Reassigning names and definitions}
\label{sec:method}

On ordinary questions the name and the definition agree, so accuracy cannot tell which one the
model follows. We therefore change which definition is bound to which name and keep everything else
fixed.

\paragraph{Reassignment.} Each binary question has two definitions: $d_{\mathrm{no}}$ states that the
condition in the instructions does not hold, and $d_{\mathrm{yes}}$ that it does (e.g.\ ``No human
attention is warranted.'' / ``A human should inspect this run.''). For option names $(n_0,n_1)$, the
\emph{aligned} arm binds $n_0$ to $d_{\mathrm{no}}$ and $n_1$ to $d_{\mathrm{yes}}$, as in the data;
the \emph{reassigned} arm binds $n_0$ to $d_{\mathrm{yes}}$ and $n_1$ to $d_{\mathrm{no}}$. Nothing
else changes. The new binding is written out in the input, and the gold answer follows the
definition. A model that reads the definitions is unaffected by the reassignment; a model that reads the
names inverts its answer.

\paragraph{Option names.} We use five polar pairs: \opt{yes}/\opt{no} and \opt{false}/\opt{true},
which Laya emitted during training, and \opt{absent}/\opt{present}, \opt{negative}/\opt{positive}
and \opt{rejected}/\opt{accepted}, which it did not. The neutral pairs \opt{0}/\opt{1} and
\opt{A}/\opt{B} are the control: for them the reassignment only reorders the definitions, so any change
measures the operation itself. We report every effect as a difference against them.
Sections~\ref{sec:r7} and~\ref{sec:r5} extend the design to random-string names and to more than
two options.

\paragraph{Data.} We use \m{nPrim} binary questions from Typed Decisions
\citep{localllama2026typed}, covering four decision tasks with \m{nPerPred} questions each: invoice
reconciliation, agent-trace triage, security-alert classification and customer-service escalation.
Each question has its own task-specific definitions, and \m{yesRate}\% of the gold answers are yes.
Another \m{nUp} questions in the same data use a generic wording whose definitions begin with the
words ``no'' and ``yes'' themselves; there the name and the definition are confounded, so we exclude
them from the main results and report them only as an upper bound (Appendix~\ref{app:bound}).
Results for Open-Jev, and the results for Laya in Table~\ref{tab:opaque}, use all \m{nOj}
questions. The 683 questions with more than two options (Section~\ref{sec:r5}) come from the test
split of the Open-Jev data \citep{zefancai2026openjevdata}.

\paragraph{Metrics.} We measure the effect of the reassignment with the flip rate, the fraction of questions
on which the model chooses a different definition in the two arms; it needs no gold labels. We
report it as a difference against the neutral controls, with 95\% intervals from 2,000 bootstrap
resamples over states. Accuracy is measured in both arms, as balanced accuracy at a threshold of 0.5
and as AUC. The aligned arm shows how well the model answers the questions to begin with (per
decision task for Laya in Appendix~\ref{app:comp}), and AUC, which does not depend on the threshold,
separates a lost ranking from a shifted decision boundary. Jev is not deterministic, so for it we
also measure a test-retest floor: the flip rate between two identical runs of the aligned arm.

\begin{table*}[t]
\centering\small
\setlength{\tabcolsep}{3pt}
\begin{tabular}{@{}llcccccl@{}}
\toprule
 & & \multicolumn{2}{c}{balanced acc.} & \multicolumn{2}{c}{AUC} & & \\
\cmidrule(lr){3-4}\cmidrule(lr){5-6}
option names & in voc. & aligned & reassigned & aligned & reassigned & flip & difference from \textit{0/1} (pp, 95\% CI) \\
\midrule
yes/no & \checkmark & \m{balANY}\% & \m{balSNY}\% & \m{aucANY}\% & \m{aucSNY}\% & \m{flipNY}\% & $+$\m{didNYZo} \tiny{[$+$\m{ciNYZolo}, $+$\m{ciNYZohi}]} \\
false/true & \checkmark & \m{balAFT}\% & \m{balSFT}\% & \m{aucAFT}\% & \m{aucSFT}\% & \m{flipFT}\% & $+$\m{didFTZo} \tiny{[$+$\m{ciFTZolo}, $+$\m{ciFTZohi}]} \\
absent/present &  & \m{balAAP}\% & \m{balSAP}\% & \m{aucAAP}\% & \m{aucSAP}\% & \m{flipAP}\% & $+$\m{didAPZo} \tiny{[$+$\m{ciAPZolo}, $+$\m{ciAPZohi}]} \\
negative/positive &  & \m{balANP}\% & \m{balSNP}\% & \m{aucANP}\% & \m{aucSNP}\% & \m{flipNP}\% & $+$\m{didNPZo} \tiny{[$+$\m{ciNPZolo}, $+$\m{ciNPZohi}]} \\
rejected/accepted &  & \m{balARA}\% & \m{balSRA}\% & \m{aucARA}\% & \m{aucSRA}\% & \m{flipRA}\% & $+$\m{didRAZo} \tiny{[$+$\m{ciRAZolo}, $+$\m{ciRAZohi}]} \\
\midrule
\textit{0/1} &  & \m{balAZO}\% & \m{balSZO}\% & \m{aucAZO}\% & \m{aucSZO}\% & \m{flipZO}\% &  \\
\textit{A/B} &  & \m{balAAB}\% & \m{balSAB}\% & \m{aucAAB}\% & \m{aucSAB}\% & \m{flipAB}\% &  \\
\bottomrule
\end{tabular}

\caption{Laya on the \m{nPrim} questions. Reassignment changes only which definition is bound to which
option name, and the gold answer follows the definition. Polar names (top) lose the decision;
neutral names (bottom) do not. \emph{in voc.} marks the two pairs Laya emitted during training.
AUC below 50\% means the ranking is reversed.}
\label{tab:main}
\end{table*}

\begin{figure*}[t]
\centering
\includegraphics[width=\textwidth]{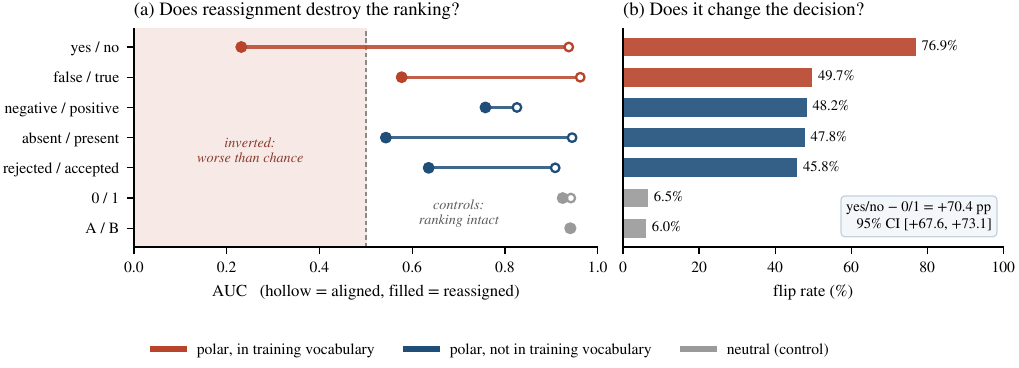}
\caption{Laya. (a) AUC in the aligned (hollow) and reassigned (filled) arms. The polar pairs fall
below chance (shaded): the ranking is reversed, not lost. (b) Flip rate for the same rows. The
neutral pairs undergo the same reassignment.}
\label{fig:diss}
\end{figure*}

\section{Results}

All experiments apply the reassignment of Section~\ref{sec:method}, and the type-error rate is 0\%
in every condition. Sections~\ref{sec:r1} and~\ref{sec:r2} show that option names override the
definitions, Section~\ref{sec:r4} that polarity contributes to this effect, and
Sections~\ref{sec:r5} and~\ref{sec:r6} that it extends to non-binary decisions and to other model
families.

\subsection{Polar names override the definitions}
\label{sec:r1}
\label{sec:r3}

Table~\ref{tab:main} shows the result on Laya. When the definitions of \opt{yes}/\opt{no} are
reassigned, 76.9\% of the decisions flip, compared with 6.5\% for \opt{0}/\opt{1} and 6.0\% for
\opt{A}/\opt{B}. The flip rate for \opt{yes}/\opt{no} is thus 70.4~pp higher than for the
\opt{0}/\opt{1} control (95\% CI [67.6, 73.1]) and 70.9~pp higher than for \opt{A}/\opt{B} (95\% CI
[68.2, 73.5]). For every polar pair it is at least 39.3~pp higher than for \opt{0}/\opt{1}.
Balanced accuracy for \opt{yes}/\opt{no} falls from 87.2\% to 28.4\%, whereas for \opt{0}/\opt{1}
it stays at 83.7\% and 83.2\%.

The gap holds across all four decision tasks (Table~\ref{tab:pertemplate}). For \opt{yes}/\opt{no}
the flip rate is at least 56.7\% in every decision task, at least 7.4 times that of
\opt{0}/\opt{1}, which never exceeds 11.3\%. The other polar pairs vary more across decision tasks,
from 12.3\% to 92.0\%; \opt{negative}/\opt{positive}, for example, flips 90.3\% of the invoice
decisions but 13.0\% of the security decisions.

\begin{table}[t]
\centering\small
\setlength{\tabcolsep}{3pt}
\begin{tabular}{@{}lccccc@{}}
\toprule
option names & invoice & agent & security & customer & min \\
\midrule
yes/no & 92.7 & 83.3 & 75.0 & 56.7 & \textbf{56.7} \\
false/true & 63.0 & 39.3 & 29.0 & 67.3 & \textbf{29.0} \\
absent/present & 56.0 & 76.7 & 39.7 & 18.7 & \textbf{18.7} \\
negative/positive & 90.3 & 31.7 & 13.0 & 58.0 & \textbf{13.0} \\
rejected/accepted & 92.0 & 37.0 & 12.3 & 41.7 & \textbf{12.3} \\
\midrule
\textit{0/1} & 2.7 & 11.3 & 6.3 & 5.7 & \textbf{2.7} \\
\bottomrule
\end{tabular}

\caption{Flip rate (\%) of Laya by decision task.}
\label{tab:pertemplate}
\end{table}

\subsection{The ranking is reversed}
\label{sec:r2}

A drop in balanced accuracy can mean that the model has become uncertain, or that it ranks the
decisions in the wrong direction. AUC distinguishes the two (Figure~\ref{fig:diss}a). For
\opt{yes}/\opt{no}, reassignment lowers Laya's AUC from 93.8\% to 23.2\%, below chance: the scores
still separate the two classes, but in the opposite direction. Because AUC does not depend on the
decision threshold, this error cannot be corrected by adjusting the threshold.

Flip rate and AUC can therefore disagree. For \opt{negative}/\opt{positive}, 48.3\% of the decisions
flip, yet AUC falls only from 82.6\% to 75.8\%: reassignment changes many decisions but leaves most
of the ranking intact. Among the polar pairs, only \opt{yes}/\opt{no} falls below chance.

\subsection{Polarity, amplified by familiarity}
\label{sec:r4}
\label{sec:r7}

The five polar pairs differ in whether Laya emitted them during training
(Section~\ref{sec:method}), which separates two contributions (Table~\ref{tab:decomp}). Moving from
neutral names to the three polar pairs that Laya never emitted raises the flip rate by 41.0~pp;
moving to the two pairs it did emit adds a further 16.0~pp. Polarity accounts for the larger part,
and it acts on names that Laya never produced during training.

\begin{table}[t]
\centering\small
\setlength{\tabcolsep}{3pt}
\begin{tabular}{@{}lcc@{}}
\toprule
option names & flip & range \\
\midrule
neutral (\textit{0/1}, \textit{A/B}) & \m{decNeut}\% & \m{decNeutLo}--\m{decNeutHi}\% \\
polar, unseen in voc.\ (3 pairs) & \m{decPolar}\% & \m{decPolarLo}--\m{decPolarHi}\% \\
polar, in voc.\ (2 pairs) & \m{decVocab}\% & \m{decVocabLo}--\m{decVocabHi}\% \\
\midrule
\multicolumn{2}{@{}l}{polarity, over neutral} & $+$\m{incrPolar}~pp \\
\multicolumn{2}{@{}l}{training-vocabulary familiarity, on top} & $+$\m{incrVocab}~pp \\
\bottomrule
\end{tabular}

\caption{Flip rate of Laya by name class. Ranges are over the four decision tasks.}
\label{tab:decomp}
\end{table}

To test whether the names' content matters beyond polarity, we name each option with a random
five-character string of letters and digits, such as \opt{xg6a6}/\opt{e97ce}. No string is a word,
neither string of a pair is a prefix of the other, and the two have no ordinal or alphabetical
relation. We draw 15 such pairs (Appendix~\ref{app:opaque}) and run all of them on Laya and
Open-Jev, and the first five on Jev. In all three models, random strings behave like the neutral
names (Table~\ref{tab:opaque}). Their flip rate differs from that of \opt{0}/\opt{1} and
\opt{A}/\opt{B} by at most 3.2~pp: 6.9\% against 7.9\% on Laya, 11.7\% against 14.8\% on Open-Jev,
and 2.0\% against 1.9\% on Jev. On the same questions, the two familiar polar pairs flip 32.2\% to
70.7\% of the decisions. The low flip rate does not come from a loss of accuracy: balanced accuracy
before reassignment differs from that with neutral names by at most 3.3~pp.

\begin{figure*}[t]
\centering
\includegraphics[width=\textwidth]{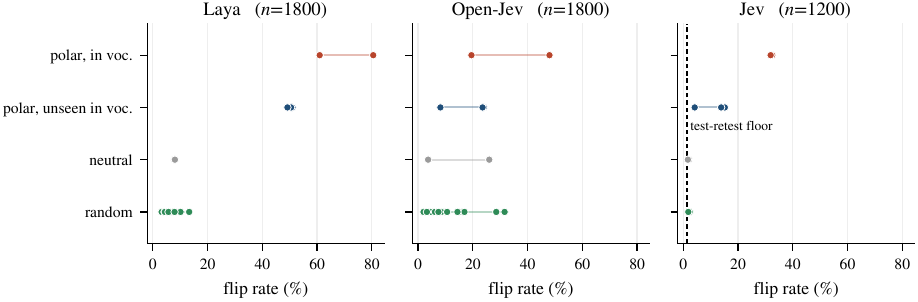}
\caption{Flip rate by name class, one dot per name pair: 15 random-string draws for Laya and
Open-Jev, five for Jev. The dashed line is Jev's test-retest floor (Section~\ref{sec:r6}).}
\label{fig:opaque}
\end{figure*}

Individual strings vary more on Open-Jev, from 2.0\% to 31.6\% across the 15 draws, than on Laya
(3.2\%--13.3\%) and Jev (1.7\%--2.4\%) (Figure~\ref{fig:opaque}).

\begin{table}[t]
\centering\small
\setlength{\tabcolsep}{2pt}
\begin{tabular}{@{}lcccc@{}}
\toprule
 & \multicolumn{2}{c}{polar} & \multicolumn{2}{c}{no polarity} \\
\cmidrule(lr){2-3}\cmidrule(lr){4-5}
model & in voc. & unseen in voc. & neutral & random \\
\midrule
\multicolumn{5}{@{}l}{\emph{flip rate}} \\
Laya & \m{opqFlipMkVocab}\% & \m{opqFlipMkPolar}\% & \m{opqFlipMkNeut}\% & \m{opqFlipMkOpq}\% \\
Open-Jev & \m{opqFlipSpVocab}\% & \m{opqFlipSpPolar}\% & \m{opqFlipSpNeut}\% & \m{opqFlipSpOpq}\% \\
Jev & \m{opqFlipApiVocab}\% & \m{opqFlipApiPolar}\% & \m{opqFlipApiNeut}\% & \m{opqFlipApiOpq}\% \\
\addlinespace
\multicolumn{5}{@{}l}{\emph{balanced accuracy, aligned arm}} \\
Laya & \m{opqBalMkVocab}\% & \m{opqBalMkPolar}\% & \m{opqBalMkNeut}\% & \m{opqBalMkOpq}\% \\
Open-Jev & \m{opqBalSpVocab}\% & \m{opqBalSpPolar}\% & \m{opqBalSpNeut}\% & \m{opqBalSpOpq}\% \\
Jev & \m{opqBalApiVocab}\% & \m{opqBalApiPolar}\% & \m{opqBalApiNeut}\% & \m{opqBalApiOpq}\% \\
\bottomrule
\end{tabular}

\caption{Random strings against the other name classes. Random-string values are means over 15
draws (five for Jev); the other columns are means over the pairs of Table~\ref{tab:main}.}
\label{tab:opaque}
\end{table}

\subsection{Beyond binary decisions}
\label{sec:r5}

We run Laya on 683 questions with three to sixteen options from the test split of the Open-Jev
data. The definitions stay unchanged and only the names change, in three ways: neutral letters
(\opt{A}, \opt{B}, \opt{C}, \dots), numbers (\opt{option 1}, \opt{option 2}, \dots), and the
original names reassigned one step, so that each name is attached to the definition of its
neighbour. The gold answer follows the definition.

Even neutral letters change 52.4\% of the decisions and lower accuracy from 56.4\% to 27.8\%;
numbers change 55.8\%. Reassigning the original names changes 79.7\% and lowers accuracy to 15.5\%
(Table~\ref{tab:e3}). Appendix~\ref{app:multiway} gives an example.

\begin{table}[t]
\centering\small
\setlength{\tabcolsep}{3pt}
\begin{tabular}{@{}lcc@{}}
\toprule
option names & acc. & changed \\
\midrule
original & \m{kAccOrig}\% & --- \\
letters \textit{A, B, C} & \m{kAccNeut}\% & \m{kChgNeut}\% \\
numbers \textit{option 1} & \m{kAccNum}\% & \m{kChgNum}\% \\
reassigned one step & \textbf{\m{kAccRot}\%} & \textbf{\m{kChgRot}\%} \\
\bottomrule
\end{tabular}

\caption{Laya on questions with more than two options. The definitions are unchanged, and the gold
answer follows the definition.}
\label{tab:e3}
\end{table}

\subsection{Other model families}
\label{sec:r6}

The effect recurs in Jev and Open-Jev (Table~\ref{tab:families}). In all three models, neutral names
flip the fewest decisions, the three polar pairs that Laya never emitted more, and the two it
emitted the most. On Jev, evaluated on the same 1,200 questions as Laya, \opt{yes}/\opt{no} flips
32.5\% of the decisions, 30.4~pp more than \opt{0}/\opt{1} (95\% CI [27.6, 33.3]); balanced
accuracy falls from 71.3\% to 51.6\%, and AUC from 81.5\% to 58.1\%. Because Jev is not
deterministic, we run the aligned arm twice on 300 questions. At most 1.33\% of the decisions change
between the two runs, so the flip rate for \opt{yes}/\opt{no} is 24 times this floor, while the
neutral pairs stay close to it. On Open-Jev, \opt{yes}/\opt{no} flips 19.5\% of the decisions,
15.9~pp more than \opt{0}/\opt{1}.

The size of the effect differs across models: for \opt{yes}/\opt{no}, Laya flips 2.4 times as often
as Jev and 4.1 times as often as Open-Jev. One possible reason for the difference between the two
open-weight models is how they score an option (Section~\ref{sec:lemma}). Laya scores a marker token
placed at the option, whereas Open-Jev averages over all tokens of the option's text, so the
one-token name is averaged with a much longer definition.

\begin{table}[t]
\centering\small
\setlength{\tabcolsep}{3pt}
\begin{tabular}{@{}lccc@{}}
\toprule
option names & Laya & Open-Jev & Jev \\
\midrule
yes/no & \m{flipNY}\% & \m{ojFlipNY}\% & \m{apiFlipNY}\% \\
false/true & \m{flipFT}\% & \m{ojFlipFT}\% & \m{apiFlipFT}\% \\
absent/present & \m{flipAP}\% & \m{ojFlipAP}\% & \m{apiFlipAP}\% \\
negative/positive & \m{flipNP}\% & \m{ojFlipNP}\% & \m{apiFlipNP}\% \\
rejected/accepted & \m{flipRA}\% & \m{ojFlipRA}\% & \m{apiFlipRA}\% \\
\textit{0/1} & \m{flipZO}\% & \m{ojFlipZO}\% & \m{apiFlipZO}\% \\
\textit{A/B} & \m{flipAB}\% & \m{ojFlipAB}\% & \m{apiFlipAB}\% \\
\midrule
polar, in voc.\ (2 pairs) & \m{decVocab}\% & \m{ojDecVocab}\% & \m{apiDecVocab}\% \\
polar, unseen in voc.\ (3 pairs) & \m{decPolar}\% & \m{ojDecPolar}\% & \m{apiDecPolar}\% \\
neutral (\textit{0/1}, \textit{A/B}) & \m{decNeut}\% & \m{ojDecNeut}\% & \m{apiDecNeut}\% \\
\midrule
questions & \m{nPrim} & \m{nOj} & \m{nApi} \\
\bottomrule
\end{tabular}

\caption{Flip rate (\%) by name pair (top) and name class (bottom) for the three models. The
ordering of the name classes holds in all three; the magnitudes differ.}
\label{tab:families}
\end{table}

\section{Related work}

\citet{kumar2025anchors} is the closest result: across eight tasks and eight 1--12B decoders,
inverted in-context demonstrations produce a semantic override rate of exactly zero. We find the
same anchoring in typed decision models, with three differences. The reassignment here is written
out in the input rather than implied by demonstrations, so the model does not have to infer it. The
output is a softmax restricted to the declared options, so the failure is invisible to a type check.
And because the gold answer follows the definition, following the name shows up as an AUC of
\m{aucSNY}\%: the ranking is reversed, not merely unchanged. \citet{le2026schemakey} treats schema \emph{keys} as an
instruction channel under constrained decoding in decoder LLMs and reports accuracy deltas; we vary
option \emph{names} in typed decision models, use neutral names as a control for the reassignment
itself, and observe a reversed ranking rather than a drop in accuracy. \citet{lim2026rubric} names our
failure mode from the other side --- safety judging as a rubric-following problem, with judges
brittle under rubric variation --- and proposes the curriculum fix we do not attempt: they vary the
rubric while holding names fixed, we hold the definitions fixed and vary which name each is bound
to. Work on sensitivity to label words and prompt format \citep{sclar2024formatspread,zheng2024mcq,liusie2023wordbias,wei2023flipped}
predicts the direction of our effect; \citet{badhe2026silentvote} analyses the renormalization step
these models perform. An independent evaluation harness reports option-\emph{order} sensitivity in
Laya \citep{ballot2026}, a complementary axis to the one studied here.

\section{Discussion}

\paragraph{What to report next to a type-error rate.} A 0\% type-error rate follows from how the
output is constructed (Section~\ref{sec:lemma}), and presenting it as a reliability number invites exactly the inference it cannot support.
It should be accompanied by a name-invariance number. The flip rate under reassignment, measured
against neutral names, is a practical candidate: it requires no gold labels, costs two extra forward
passes per question, and on Laya it reveals a flip rate of 76.9\% for \opt{yes}/\opt{no} that
accuracy with the original names does not show.

\paragraph{For practitioners.} Two mitigations follow directly. Use neutral option names and carry the
meaning in the definitions, which on Laya gives a flip rate of 6.5\% instead of 76.9\%; or train
the model to follow the definitions by randomizing option names, which is cheap for a model of
Laya's size. Curriculum-based rubric-following \citep{lim2026rubric},
option-ID debiasing \citep{zheng2024mcq} and word-bias correction \citep{liusie2023wordbias} are
available mitigations we do not evaluate here.

\section{Limitations}

We study two open-weight models, which differ in how they score an option, on English questions
only; whether the explanation by scoring (Section~\ref{sec:r6}) holds for other architectures is
untested. Open-Jev also answers less accurately
before the reassignment: its balanced accuracy is 59.8\%--63.9\% across name classes, against
79.4\%--89.6\% for Laya (Table~\ref{tab:opaque}), so its lower flip rate may also reflect a
preference for one definition regardless of the state. Jev is a closed-source
model accessed through an API at one point in time: we observe only its returned distribution, it
is not deterministic --- hence the test-retest floor of Section~\ref{sec:r6} --- and the model served
under that name may change. Name pairs differ in how far they preserve meaning, which is why
Section~\ref{sec:r1} reports the minimum over decision tasks rather than the mean; a human
annotation of meaning preservation across name pairs would sharpen the middle rows of
Table~\ref{tab:decomp} and is the natural next step. On the questions with more than two options, accuracy with
the original names is only 56.4\%, so we treat that result as directional. We report the vulnerability and two mitigations but
evaluate neither.

\section{Conclusion}
Type safety guarantees that a typed decision model returns an option from the declared set, but it does not guarantee that the model interprets those options according to their explicit definitions. Holding the question, state, option names, and definition texts fixed, we find that changing only which definition is bound to which name can substantially alter model decisions. For \opt{yes}/\opt{no}, the flip rate under this reassignment is \m{hlDid}~pp higher than under the neutral \opt{0}/\opt{1} control, and AUC falls from \m{aucANY}\% to \m{aucSNY}\%. These results show that structurally valid decisions can remain strongly dependent on the semantic cues carried by option names. Type safety guarantees where a decision can land, but not what the options mean to the model.

\bibliography{refs}

@article{kumar2025anchors,
  title   = {Semantic Anchors in In-Context Learning: Why Small {LLMs} Cannot Flip Their Labels},
  author  = {Krishna Kumar, Anantha Padmanaban},
  journal = {arXiv preprint arXiv:2511.21038},
  year    = {2025}
}

@article{lim2026rubric,
  title   = {Reliable to Expressive: A Curriculum for Rubric-Following Safety Judges},
  author  = {Lim, Yongtaek and Choi, Hyeji and Kim, Minwoo},
  journal = {arXiv preprint arXiv:2606.09165},
  year    = {2026},
  note    = {{ICML} 2026 Workshop on AIWILDS}
}

@inproceedings{le2026schemakey,
  title     = {Schema-Key Wording as an Instruction Channel in Structured Generation under Constrained Decoding},
  author    = {Le, Yifan},
  booktitle = {Proceedings of AACL-IJCNLP},
  year      = {2026},
  note      = {arXiv:2604.14862}
}

@article{li2026json,
  title   = {When {JSON} Is Not Enough: Semantic Reliability of Schema-Constrained {LLM} Ordering Agents},
  author  = {Li, Yin},
  journal = {arXiv preprint arXiv:2607.18261},
  year    = {2026}
}

@article{singh2026sob,
  title   = {The Structured Output Benchmark: A Multi-Source Benchmark for Evaluating Structured Output Quality in Large Language Models},
  author  = {Singh, Abhinav Kumar and Khurdula, Harsha Vardhan and Khemlani, Yoeven D. and Agarwal, Vineet},
  journal = {arXiv preprint arXiv:2604.25359},
  year    = {2026}
}

@article{usman2026phantomfill,
  title   = {{PhantomFill}: When the Form Demands an Answer, Language Models Invent One},
  author  = {Usman, Rana Muhammad},
  journal = {arXiv preprint arXiv:2607.20492},
  year    = {2026}
}

@article{yourprompt2026,
  title   = {Your Prompt Is Not the Only Prompt: How Much Do {LLMs} Weight Structured-Output Schema Descriptions?},
  author  = {Lin, Sin-Ying},
  journal = {arXiv preprint arXiv:2608.08254},
  year    = {2026}
}

@article{badhe2026silentvote,
  title   = {The Silent Vote: Improving Zero-Shot {LLM} Reliability by Aggregating Semantic Neighborhoods},
  author  = {Badhe, Sanket and Tiwari, Priyanka and Shah, Deep},
  journal = {arXiv preprint arXiv:2605.09739},
  year    = {2026},
  note    = {{GEM} Workshop at {ACL} 2026}
}

@inproceedings{zheng2024mcq,
  title     = {Large Language Models Are Not Robust Multiple Choice Selectors},
  author    = {Zheng, Chujie and Zhou, Hao and Meng, Fandong and Zhou, Jie and Huang, Minlie},
  booktitle = {International Conference on Learning Representations (ICLR)},
  year      = {2024},
  note      = {arXiv:2309.03882}
}

@article{liusie2023wordbias,
  title   = {Mitigating Word Bias in Zero-shot Prompt-based Classifiers},
  author  = {Liusie, Adian and Manakul, Potsawee and Gales, Mark},
  journal = {arXiv preprint arXiv:2309.04992},
  year    = {2023}
}

@inproceedings{sclar2024formatspread,
  title     = {Quantifying Language Models' Sensitivity to Spurious Features in Prompt Design},
  author    = {Sclar, Melanie and Choi, Yejin and Tsvetkov, Yulia and Suhr, Alane},
  booktitle = {International Conference on Learning Representations (ICLR)},
  year      = {2024},
  note      = {arXiv:2310.11324}
}

@article{wei2023flipped,
  title   = {Larger Language Models Do In-Context Learning Differently},
  author  = {Wei, Jerry and Wei, Jason and Tay, Yi and Tran, Dustin and Webson, Albert and Lu, Yifeng and Chen, Xinyun and Liu, Hanxiao and Huang, Da and Zhou, Denny and Ma, Tengyu},
  journal = {arXiv preprint arXiv:2303.03846},
  year    = {2023}
}

@misc{typesafe,
  title  = {Jev documentation},
  author = {{TypeSafe}},
  year   = {n.d.},
  howpublished = {\url{https://docs.typesafe.ai}; Introducing System One models and Jev, \url{https://typesafe.ai/blog/introducing-system-one-models-and-jev}},
  note   = {Accessed 2026-09-22}
}

@misc{convai2026laya,
  title  = {laya-typed-decisions},
  author = {{Convai Innovations}},
  year   = {n.d.},
  howpublished = {Hugging Face, \url{https://huggingface.co/convaiinnovations/laya-typed-decisions}}
}

@misc{kotoba2026openjev,
  title  = {open-jev-deberta-v3-large},
  author = {{com-kotobalabs}},
  year   = {n.d.},
  howpublished = {Hugging Face, \url{https://huggingface.co/com-kotobalabs/open-jev-deberta-v3-large}; code, \url{https://github.com/kotoba-lang/typed-decisions}}
}

@misc{localllama2026typed,
  title  = {Typed Decisions},
  author = {{LocalLLaMA}},
  year   = {n.d.},
  howpublished = {Hugging Face, \url{https://huggingface.co/datasets/LocalLLaMA/typed-decisions}}
}

@article{warner2024modernbert,
  title   = {Smarter, Better, Faster, Longer: A Modern Bidirectional Encoder for Fast, Memory Efficient, and Long Context Finetuning and Inference},
  author  = {Warner, Benjamin and Chaffin, Antoine and Clavi{\'e}, Benjamin and Weller, Orion and Hallstr{\"o}m, Oskar and Taghadouini, Said and Gallagher, Alexis and Biswas, Raja and Ladhak, Faisal and Aarsen, Tom and Cooper, Nathan and Adams, Griffin and Howard, Jeremy and Poli, Iacopo},
  journal = {arXiv preprint arXiv:2412.13663},
  year    = {2024}
}

@inproceedings{he2023debertav3,
  title     = {{DeBERTaV3}: Improving {DeBERTa} using {ELECTRA}-Style Pre-Training with Gradient-Disentangled Embedding Sharing},
  author    = {He, Pengcheng and Gao, Jianfeng and Chen, Weizhu},
  booktitle = {International Conference on Learning Representations (ICLR)},
  year      = {2023},
  note      = {arXiv:2111.09543}
}

@misc{ballot2026,
  title  = {Option-order sensitivity in a typed decision head},
  author = {{Vignesh Labs}},
  year   = {2026},
  note   = {ballot evaluation harness, option-order flip rate 0.433}
}

@misc{zefancai2026openjevdata,
  title  = {Open-Jev: typed decision datasets},
  author = {{ZefanCai}},
  year   = {n.d.},
  howpublished = {Hugging Face, \url{https://huggingface.co/datasets/ZefanCai/Open-Jev}; code, \url{https://github.com/Zefan-Cai/Open-Jev-Dev}}
}

\appendix

\section{Accuracy of the aligned arm}
\label{app:comp}
\begin{table}[h]
\centering\footnotesize
\setlength{\tabcolsep}{3pt}
\begin{tabular}{@{}lp{.37\columnwidth}ccc@{}}
\toprule
decision task & instructions & $n$ & bal.\ acc. & AUC \\
\midrule
invoice & \textit{``The invoice reconciles with the purchase order and the recorded delivery.''} & 300 & 87.8\% & 97.1\% \\
agent & \textit{``This trace requires human review.''} & 300 & 87.4\% & 94.7\% \\
security & \textit{``This alert reflects genuinely malicious or unauthorised activity.''} & 300 & 93.1\% & 98.7\% \\
customer & \textit{``This conversation requires a human agent rather than automated handling.''} & 300 & 78.6\% & 91.4\% \\
\bottomrule
\end{tabular}

\caption{Laya, aligned arm, \opt{yes}/\opt{no}, by decision task.}
\label{tab:comp}
\end{table}

\section{Flip rate by decision task}
\begin{figure}[h]
\centering
\includegraphics[width=\linewidth]{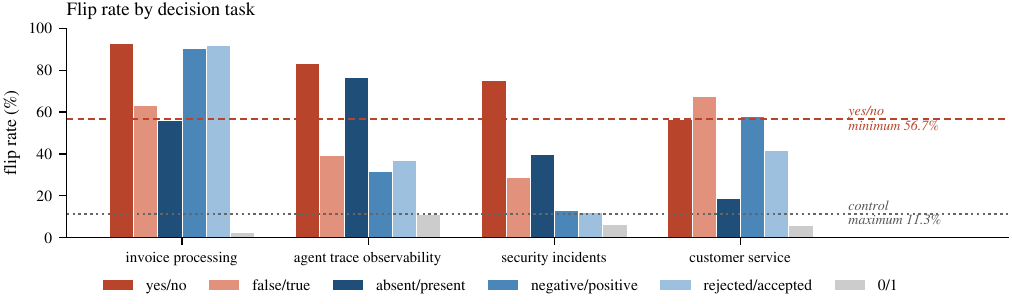}
\caption{Flip rate of Laya by decision task and name pair.}
\label{fig:pt}
\end{figure}

\begin{figure}[h]
\centering
\includegraphics[width=.86\linewidth]{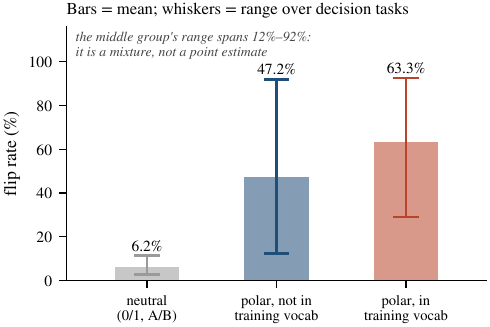}
\caption{Flip rate of Laya by name class. Bars are means over pairs; whiskers span decision tasks.}
\label{fig:dec}
\end{figure}

Figure~\ref{fig:pt} plots Table~\ref{tab:pertemplate}; Figure~\ref{fig:dec} plots
Table~\ref{tab:decomp}, with the range over decision tasks as whiskers.

\section{Questions with generic definitions}
\label{app:bound}
On the \m{nUp} questions excluded in Section~\ref{sec:method}, whose definitions begin with the
words ``no'' and ``yes'', reassignment flips 87.7\% of the decisions for \opt{yes}/\opt{no}, 77.2~pp
more than for \opt{0}/\opt{1} (95\% CI [73.0, 81.3]). AUC falls to 8.6\% for \opt{yes}/\opt{no} and
3.0\% for \opt{false}/\opt{true}. Because the name and the definition are confounded on these
questions, we report this only as an upper bound.

\section{Random-string names}
\label{app:opaque}
Table~\ref{tab:opaquenames} lists the 15 pairs of Section~\ref{sec:r7}. They come from a seeded
generator whose acceptance rule depends only on the names already accepted, so drawing more pairs
does not change the earlier ones. The first row lists the five pairs run on Jev.
\begin{table}[h]
\centering\scriptsize
\setlength{\tabcolsep}{2pt}
\begin{tabular}{@{}lllll@{}}
\toprule
\opt{xg6a6}/\opt{e97ce} & \opt{30d6n}/\opt{j87g8} & \opt{hh0t5}/\opt{blj8b} & \opt{94wwf}/\opt{dv7vq} & \opt{6nfxk}/\opt{7qnyc} \\
\midrule
\opt{gc46d}/\opt{wgh8i} & \opt{18gc6}/\opt{idqj2} & \opt{fbg13}/\opt{cz922} & \opt{2cdod}/\opt{lh8uo} & \opt{yn6by}/\opt{syji6} \\
\opt{k63g1}/\opt{za2ml} & \opt{mq5h1}/\opt{nvoz2} & \opt{v5mai}/\opt{paw1t} & \opt{u5r4o}/\opt{rw9lp} & \opt{0se36}/\opt{4r2qe} \\
\bottomrule
\end{tabular}

\caption{The 15 random-string name pairs.}
\label{tab:opaquenames}
\end{table}

\section{An example with more than two options}
\label{app:multiway}
The question below is from the test split of the Open-Jev data (Section~\ref{sec:r5}). The
instructions ask ``What exact integer is stored in result after this program finishes?'', and the
state is the program
\begin{verbatim}
x = 8
for item in [6, 9, -9, 1]:
    if item > 1:
        x += item
    else:
        x -= 5
result = x * 0
\end{verbatim}
The answer is 0. Table~\ref{tab:mwexample} gives Laya's probabilities for the four definitions. With
the original names, Laya chooses ``The exact integer result is 0'' (50.4\%). When the names are
reassigned one step, the name \opt{0} is attached to ``The exact integer result is -17'', and Laya
chooses that option (41.2\%); the correct definition, now named \opt{5}, receives 19.9\%.
\begin{table}[h]
\centering\footnotesize
\setlength{\tabcolsep}{3pt}
\begin{tabular}{@{}lcccc@{}}
\toprule
 & \multicolumn{2}{c}{original} & \multicolumn{2}{c}{reassigned} \\
\cmidrule(lr){2-3}\cmidrule(lr){4-5}
definition & name & $p$ (\%) & name & $p$ (\%) \\
\midrule
The exact integer result is 5   & \opt{5}   & 19.4 & \opt{-3}  & 18.3 \\
The exact integer result is -3  & \opt{-3}  & 15.6 & \opt{-17} & 20.5 \\
The exact integer result is -17 & \opt{-17} & 14.5 & \opt{0}   & \textbf{41.2} \\
The exact integer result is 0   & \opt{0}   & \textbf{50.4} & \opt{5} & 19.9 \\
\bottomrule
\end{tabular}
\caption{Laya on one question with four options. Each row is one definition; the name columns give
the name attached to it in each arm.}
\label{tab:mwexample}
\end{table}

\end{document}